%% file: 0_main.tex
\documentclass[letterpaper, 10 pt, conference]{ieeeconf}  

\IEEEoverridecommandlockouts                              

\usepackage{lipsum} 
\usepackage{epsfig}
\usepackage{float}
\usepackage{bbold}
\usepackage{bm}
\usepackage{hyperref}
\usepackage{todonotes}

\usepackage[noadjust]{cite}

\usepackage{color,soul} 

\usepackage{amsmath}

\title{\LARGE \bf
Placeit! A Framework for Learning Robot Object Placement Skills
}

\begin{document}

\author{Amina Ferrad$^{*1}$, Johann Huber$^{*1}$,
François Hélénon$^{1}$, Julien Gleyze$^{1}$, Mahdi Khoramshahi$^{1}$, and Stéphane Doncieux$^{1}$ 
\thanks{$^*$ equal contribution and corresponding authors}
\thanks{$^{1}$Sorbonne Université, CNRS, Institut des Systèmes Intelligents et de Robotique, ISIR, F-75005 Paris, France {\tt\small \{ferrad, huber, helenon, gleyze, khoramshahi, doncieux\}@isir.upmc.fr}}}

\maketitle
\thispagestyle{empty}
\pagestyle{empty}




\input{tex_files/abstract}

\input{tex_files/1_introduction}

\input{tex_files/2_related_works}

\input{tex_files/3_method}

\input{tex_files/4_experiments}

\input{tex_files/5_results_and_discussion}

\input{tex_files/6_conclusions}


\input{tex_files/acknowledgment}



\bibliographystyle{IEEEtran}

\input{tex_files/bilbio}
\input{tex_files/appendices}


\end{document}

%% file: tex_files/abstract.tex

\begin{abstract}

Robotics research has made significant strides in learning, yet mastering basic skills like object placement remains a fundamental challenge. A key bottleneck is the acquisition of large-scale, high-quality data, which is often a manual and laborious process. Inspired by \textit{Graspit!}, a foundational work that used simulation to automatically generate dexterous grasp poses, we introduce \textit{Placeit!}, an evolutionary-computation framework for generating valid placement positions for rigid objects. Placeit! is highly versatile, supporting tasks from placing objects on tables to stacking and inserting them. Our experiments show that by leveraging quality-diversity optimization, Placeit! significantly outperforms state-of-the-art methods across all scenarios for generating diverse valid poses. A pick\&place pipeline built on our framework achieved a 90\% success rate over 120 real-world deployments. This work positions Placeit! as a powerful tool for open-environment pick-and-place tasks and as a valuable engine for generating the data needed to train simulation-based foundation models in robotics.

\end{abstract}

%% file: tex_files/1_introduction.tex
\section{INTRODUCTION}
\label{sec:1_introduction}

Despite decades of effort in robot learning, basic manipulation skills like grasping and placing objects remain a significant challenge \cite{zhang2025robustdexgrasp, padalkar2023openxembodiement, zhao2025anyplace, jermann2024efficient, cao2021suctionnet}. As the field has evolved, data acquisition has grown to be a primary consideration for the community \cite{huber2023quality, eppner2021acronym, zhao2025anyplace, noh2024learning, simeonov2023shelving}. This is mostly due to the prevailing belief that major breakthroughs, which one could refer to as the "ChatGPT moment" of robotics, will arise from the availability of large-scale data \cite{fizoori2025foundation}.

This has led to two primary data acquisition paradigms. The first involves collecting massive amounts of high-quality data from the physical world \cite{padalkar2023openxembodiement, kim2024openvla, khazatsky2024droid, barreiros2025careful, walke2023bridgedata}. This approach provides realistic information but the data collection process is slow and hardly scalable. The second paradigm utilizes simulation to accelerate data collection and safely train robot control models \cite{zhang2025robustdexgrasp, lum2024dextrah, sum2024sim2real}. Recent research often combines both, either to build foundational models \cite{bjorck2025gr00t} or to enhance simulation fidelity through methods like real2sim2real \cite{patel2025real}.

Simulators should allow to automatically generate data for robotic learning, and scale-up the process in a safer and more efficient manner than in the physical world. However, a considerable amount of work also requires manually collected data in simulation \cite{bjorck2025gr00t, hussing2023robotic}: the automatic synthesis of meaningful training data remains a significant challenge. Although there has been extensive research on automated grasping \cite{huber2023quality, eppner2021acronym, huber2023domain}, other manipulation tasks, such as placing, have not been as thoroughly investigated.

The automatic generation of data for robotics was strongly inspired by the seminal work of \textit{Graspit!} \cite{miller2004graspit}, a framework that generates a set of dexterous grasps in simulation based on analytic criteria, given the 3D models of an object and a gripper. Inspired by this foundational work, we introduce \textit{Placeit!}, an evolutionary-computation based framework for automatically generating valid placement positions of rigid objects in a simulated environment.

This framework is versatile, taking two objects as input—one to be placed and a support object. This allows it to handle various placement scenarios, including table-top, hanging, stacking, and inserting (Fig. \ref{fig:qdg6dof_overview}). Our approach leverages Quality-Diversity (QD) optimization, a family of evolutionary algorithms that has proven effective for generating diverse and high-quality data for robot learning \cite{cully2022qd}.

The key contributions of this work are the following:

\input{tex_files/figures/qd_place_overview}

\begin{itemize}
    \item We introduce Placeit!, a versatile framework for automatically generating diverse and valid placement positions of rigid objects in various simulated scenes;
    \item We demonstrate that a state-of-the-art QD method leveraged in the Placeit! framework outperforms standard approaches in most of the considered scenarios by a large margin;
    \item We also introduce \textit{Quality-Diversity Grasp-and-Place} (QDGP), a pipeline built on the top of Placeit! to do pick\&place tasks in various scenarios. QDGP achieved about 90\% success rate over 120 deployments through the use of domain randomization to sample the most promising poses.
\end{itemize}

The code had been made publicly available\footnote{\url{https://gitlab.isir.upmc.fr/l2g/placeit}}. Visualizations of both the simulated and the real-world experiments are provided with the attached video\footnote{Released soon.}.

%% file: tex_files/figures/qd_place_overview.tex
\begin{figure}[t]
  \centering
\centering
  \includegraphics[width=0.7\columnwidth]{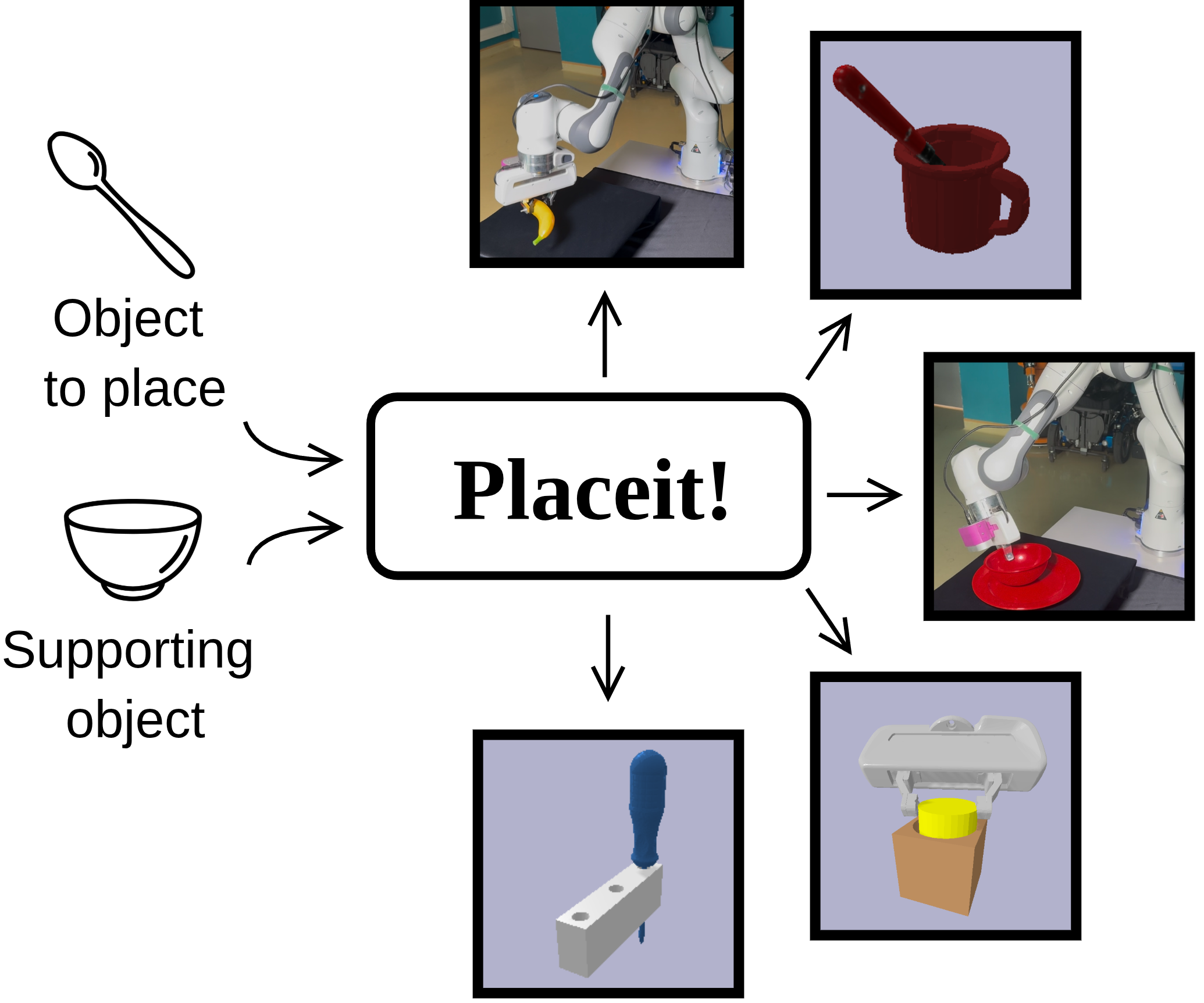}
  \caption{\textit{Placeit!} takes two object meshes as input and automatically finds diverse ways for them to interact. The resulting poses can be applied to a wide variety of scenarios, enabling robust robotic manipulations.}
  \label{fig:qdg6dof_overview}
\end{figure}

%% file: tex_files/2_related_works.tex

\section{RELATED WORKS}
\label{sec:2_related_works}


\textbf{\textit{Learning to Place in Robotics.}} Research in robot placement has long relied on manually labeled or expert-provided data. This approach involves defining placement poses programmatically for predefined objects, which may include primitive shapes \cite{garrett2018ffrob} or common household items \cite{wan2019regrasp, li2024stable, simeonov2023shelving}. Numerous works that address object placement, including those leveraging supervised learning from human demonstrations, fall into this category. Examples range from studies focused exclusively on placing \cite{shan2025slot} to those with a broader scope covering various manipulation tasks \cite{padalkar2023openxembodiement, bjorck2025gr00t}. Another common approach is the use of Reinforcement Learning (RL), which often requires carefully hand-crafted reward functions to guide the learning process \cite{pavlichenko2025dexterous} or constrains the problem to a specific, more tractable setup \cite{wolski2017hindsight}. Many pick-and-place scenarios simplify the placement problem by merely defining a target position for the end-effector, typically above a large container, before releasing the object \cite{jermann2024efficient, cao2021suctionnet, fang2023anygrasp, lum2024dextrah}. Existing methods are heavily reliant on human intervention, making data acquisition slow, tedious, and difficult to scale. This manual dependency is a significant bottleneck for generating the large-scale datasets required for modern foundation models in robotics. Our framework overcomes this limitation by automatically generating valid placement poses across diverse scenarios.


Many works on robot placement rely on analytical criteria to sample valid poses. A common approach involves aligning vertices or mesh faces of an object with a target surface in simulation. A candidate pose is typically considered valid if the projection of its center of mass onto the surface, along the gravity vector, falls within the surface's polygon \cite{lertkultanon2018certified, harada2014validating, street2024right}. However, this method has significant limitations. It is highly sensitive to the quality of the object's mesh, overlooks the dynamics of physical interactions during placement, and fails to differentiate between stable and unstable placement positions, as long as the geometric criteria are met. In contrast, our framework explores the physical interactions between the object and the support in a simulated environment. This enables us to not only validate a placement but also to quantify its stability. Furthermore, we use a domain-randomization-based approach to simulate local perturbations on the placed object, which helps in identifying stable poses versus those that are local maxima of potential energy.

Recent research has increasingly used simulations to test the physical interactions between objects and surfaces for generating placement poses. Zhao et al. \cite{zhao2025anyplace} leveraged principal component analysis to sample placement poses by aligning objects with their supports. Noh et al. uniformly sampled object orientations and simulated dropping them onto a plane to discover valid placement poses \cite{noh2024learning}. Simeonov et al. \cite{simeonov2023shelving} rely on manually coded procedures to sample initial candidates, which are then validated through simulation. While these works have made contributions to specific areas—such as automatic labeling for tabletop scenarios \cite{noh2024learning}, object alignment for stacking and inserting \cite{zhao2025anyplace}, or addressing diverse placement scenarios with human-labeled poses \cite{simeonov2023shelving}, our framework is designed to automatically sample placements across all of these scenarios. Moreover, our sampling scheme, which is based on quality-diversity optimization, significantly outperforms the automatic sampling methods used in the aforementioned papers across nearly all tested scenarios.


\textbf{\textit{Quality Diversity.}} 
Quality-Diversity methods are optimization algorithms that aim to generate a set of diverse and high-performing solutions to a given problem. An increasing amount of works are dedicated to using QD for generating data for robot learning. This includes the generation of demonstrations for locomotion \cite{mace2023quality}, grasping trajectories \cite{huber2023quality}, or object-centric grasps \cite{huber2024speeding, huber2025qdgset}, but also the production of adversarial objects to grasp \cite{morrison2020egad}. 
Application of QD algorithms for the acquisition of manipulation skills are however limited to grasping \cite{huber2023quality}. The potential of these algorithms to generate diverse data in hard exploration problems suggests that it could be leveraged for prehensile manipulation tasks. The present paper fills this gap by investigating how to leverage QD optimization to build a versatile tool for generating placement position of objects.

%% file: tex_files/3_method.tex
\input{tex_files/figures/qd_place_principle}


\section{METHOD}


\subsection{Placing}
\label{sec:III_A_placing}

Let $P \subseteq \mathrm{SE}(3)$ be the space of rigid object poses, and $P^* \subseteq P$ the space of stable poses. This section discusses how the literature generates diverse, stable object placements for robotic manipulation—that is, exploring a \textit{parameter space} $\Theta$ to find associated stable positions in $P^*$.



\textbf{\textit{Definition.}} We consider the placement of an object $o$ in interaction with a supporting object $s$. For an individual $\theta_i$, let $X_i^o(t) \in P$ be the pose of a rigid body $o$ at time $t$ in the supporting object frame that we assume static. The object $o$ is placed on $s$ if and only if for each $t\in t_{bvp}, ..., t_{evp}$:

\begin{equation}
\label{eq:place_theoretical}
    \left\{\begin{matrix}
        \sum \vec{F}_{ext}(t) = \vec{0} \\
        \sum \vec{\tau}_{ext}(t) = \vec{0} \\
        in\_contact_{o,s}(t)=1
    \end{matrix}\right. 
\end{equation}
where $\vec{F}_{ext}$ and $\vec{\tau}_{ext}$ are respectively the external forces and external torques applied on $o$ at timestep $t$, $t_{bvp}, t_{evp} \in\mathbb{N}^+$ are respectively the timestep that begins the validation and the one that ends the validation of the placement (with $t_{bvp} < t_{evp}$), and $in\_contact_{o,s}:\mathbb{N}^{+}\rightarrow \left\{0, 1 \right\}$ is a binary function that returns 1 if $o$ and $s$ are in contact at timestep $t$ and 0 otherwise. From an empirical perspective, the placement of $o$ on $s$ can be empirically verified if for each $t\in t_{bvp}, ..., t_{evp}$: 
\begin{equation}
\label{eq:place_empirical}
    \left\{\begin{matrix}
         \vec{v}_{o/s}(t) \approx \vec{0}\\
        in\_contact_{o,s}(t)=1
    \end{matrix}\right. 
\end{equation}
where $\vec{v}_{o/s}$ is the velocity (linear and angular) of $o$ in the frame of $s$.


\textbf{\textit{Sampling schemes.}} To discover placing poses, previous works essentially relied on hand-crafted search space to guide the search toward promising solutions, and then explore $P$ by randomly sampling $\Theta$. 

The existing literature on sampling placement poses primarily uses three main strategies. The first involves aligning two surfaces of the considered objects \cite{lertkultanon2018certified, harada2014validating, street2024right}. The second aligns objects based on their geometry, which is particularly effective for problems with symmetrical properties, such as peg-in-hole or stacking \cite{zhao2025anyplace}. The third strategy samples random poses for one object relative to another, such as placing an object above a table with a random orientation, to find diverse placement poses \cite{noh2024learning}. 

These methods often rely on a predefined search space combined with a brute-force search. Placeit!, moves beyond these limitations by integrating the search space with an optimized exploration strategy based on recent advances in evolutionary computation.


\subsection{Placeit!}



\textbf{\textit{Definition.}} Let $\theta \in \Theta$ be an \textit{individual}. Let $\Phi \subseteq \mathbb{R}^{n_b}$ be the \textit{feature space}, and $\zeta_{\Phi}:\Theta \rightarrow \Phi$ the \textit{feature function}, which assigns a \textit{feature descriptor} $\phi_\theta = \zeta_{\Phi}(\theta)$ to each $\theta$. The \textit{fitness function} is $f:\Theta\rightarrow \mathbb{R}$, and $d_{\Phi}:\Phi^2 \rightarrow \mathbb{R}$ is a distance function within $\Phi$. The goal is to generate an \textit{archive} $A$ such that:
\begin{equation}
\left\{\begin{matrix}
\forall \phi \in \Phi_{reach}, \, \exists \theta \in A, \, d_{\Phi}(\zeta_{\Phi}(\theta), \phi) < \epsilon \\
\forall \theta' \in A, \, \theta' =  \text{argmax}_{\theta\in N(\phi_{\theta'})}f(\theta) 
\end{matrix}\right.
\label{eq:qd_background}
\end{equation}
where $\Phi_{reach} \subseteq \Phi$ is the \textit{reachable feature space}, $\epsilon\in\mathbb{R}^{+*}$ defines the density of $\Phi_{reach}$ paving, and $N(\phi_{\theta'})= \{ \theta \mid neighbor_{d_{\Phi}}(\phi_\theta, \phi_{\theta'}) \}$ is the set of solutions with close projections in $\Phi$. $\zeta_{\Phi}$ is deterministic.

\input{tex_files/figures/qdgp_principle}



\textbf{\textit{Principle.}} The core principle of Placeit! is to use quality-diversity optimization to discover a diverse set of valid and stable placement poses (Fig. \ref{fig:qdp_principle}). The framework leverages the powerful exploration properties of evolutionary algorithms \cite{cully2022qd, huber2023quality}, making the discovery of these poses both automatic and sample-efficient. Unlike other methods that rely on priors specific to certain objects and scenarios \cite{lertkultanon2018certified, harada2014validating, street2024right, zhao2025anyplace, noh2024learning}, Placeit! is a generic framework that can be applied to any 3D objects provided as input for both the object to be placed and the supporting object.

\textbf{\textit{Evaluation.}} The process begins by randomly sampling a population of parameters $\{\theta_i\}_{i=0}^n$ from the parameter space $\Theta$. These parameters are then used to generate a set of initial poses $\{X_i^o(0)\}_{i=0}^n$ for the object $o$ relative to the supporting object $s$. The projections used to obtain $X_i^o(0)$ from $\theta_i$ are detailed in section \ref{sec:experiments}, \textit{Compared methods}. We discard any poses that result in an overlap between the two objects.

Each valid initial pose is then evaluated in a simulation. Gravity is applied to the object $o$ for a duration of $T$ time steps, allowing it to fall and settle. The result of this simulation is a sequence of poses $\{X_i^o(t)\}_{i=0}^n$ for $t \in 0, ..., T$. A placement attempt is considered successful if the object reaches a stable state at the end of the simulation. This stability is determined by the variance of the object's pose over the final $\delta_t = t_{evp} - t_{bvp}$ steps of the simulation (here $t_{evp}=T$). Specifically, a placement is successful if:
\begin{equation}
    \sigma_i^o < \sigma^{st} \hspace{1mm} | \hspace{1mm} \sigma_i^o = \sigma(\{X_i^o(t)\}_{t=T-\delta_t}^{T})
    \label{eq:valid_place}
\end{equation}
where $\sigma_i^o$ is the variance of the object's pose, and $\sigma^{st}$ is a predefined threshold that distinguishes stable poses from unstable ones.

For each successful placement, a quality score (fitness) and a descriptor (feature) are computed. The fitness of $\theta_i$ is defined as $f_{\theta_i} = -\sigma_i^o$, such that a lower variance (higher stability) corresponds to a better fitness. The feature is the final pose of the object relative to the support, $\phi_{\theta_i} = X_i^o(T)$, which describes the outcome of the placement.

\textbf{\textit{Optimization.}} The core of the Placeit! framework is its use of a QD algorithm, which maintains an archive of solutions \cite{mouret2015mapelites}. This archive acts as a memory of the exploration process, storing the highest-performing solution for each distinct feature. Solutions, defined by their initial poses and corresponding outcomes, are added to the archive based on their feature and fitness. If a cell in the archive (corresponding to a specific feature range) is empty, a new solution is added. Otherwise, if a solution with a higher fitness is found, it replaces the existing one. This process progressively populates the archive with a diverse set of high-quality solutions.

New parameters for the next iteration are sampled from this archive and mutated with random noise, allowing the algorithm to build on previously successful discoveries. The algorithm's output is the optimized archive, which contains a collection of diverse and robust final poses. The objective is to optimize the coverage of the \textit{successful feature space} $\Phi^s \subseteq \Phi$ which represents all final poses that meet the stability criteria. The QD algorithm's local competition ensures that the optimization focuses on finding the most stable poses within each feature region.

The final output of the algorithm is the archive of robust, diverse final poses $X_i^o(T)$. These poses can then be used for deployment on a physical robot. While some complex cases may require dedicated methods for the approach phase \cite{haustein2019object}, this separate problem is outside the scope of this work.

\input{tex_files/figures/qdplace_tasks}


\textbf{\textit{Domain randomization.}} The fitness function we use efficiently identifies valid placement states in a single evaluation, a common approach in QD optimization \cite{cully2022qd}. However, a rigid body's true stability is determined by its resistance to small perturbations, which relates to potential energy analysis \cite{batlle2022rigid}. In this work, we focus on identifying stable equilibria in the sense of a pose where the object remains at rest without toppling or rolling away under small perturbations. To achieve this, we introduce a domain randomization-based filter that distinguishes stable equilibria (minima of potential energy) from unstable ones (maxima).

For a given final object state, $X_i^o(T)$, we apply a perturbation $\epsilon \in \mathbb{R}^6$ to get a new state $\tilde{X}_i^o(T) = X_i^o(T) + \epsilon$. We then evaluate $\tilde{X}_i^o(T)$ by applying gravity for $\delta_{t_{DR}}$ additional steps, which gives us a set of states, $\{\tilde{X}_i^o(T+t)\}_{t=0}^{\delta_{t_{DR}}}$. A placing pose, $X_i^o(T)$, is considered stable if:
\begin{equation}
    \sigma_i^{o^{DR}} < \sigma^{DR} \hspace{1mm} | \hspace{1mm} \sigma_i^{o^{DR}} = \sigma(\{\tilde{X}_i^o(T+t)\}_{t=0}^{\delta_{t_{DR}}})
    \label{eq:dr_validation_place}
\end{equation}

Here, $\sigma_i^{o^{DR}}$ is the variance of the object's state during the $\delta_{t_{DR}}$ stabilization steps after the perturbation, and $\sigma^{DR}$ is a predefined variance threshold. Defining $\sigma^{DR}$ is challenging since it includes both positional and rotational variance. The value of $\sigma^{DR}$ is here defined empirically, making the "robust" label subjective. However, this evaluation method still allows us to determine which states are more stable than others. Our empirical results show that this is a powerful analytical tool (section \ref{sec:results_and_discussion}).

To properly assess an object's potential energy, its resistance to a range of small perturbations must be tested, as a single perturbation might not reveal an unstable state. We perform this test $M\in\mathbb{N}^{*+}$ times. A larger value of $M$ gives a better estimate of the potential energy but requires more computation.

\textbf{\textit{Quality-Diversity Grasp-and-Place.}} To explore the usage of Placeit! for pick\&place tasks, we introduce QDGP, a pick-and-place pipeline based on quality-diversity generated data (Fig. \ref{fig:qdgp_principle}). First, a set of placing poses $X_i^o(T)$ is generated for a given scenario. Simultaneously, a set of grasping poses is produced on the object $o$ and the gripper using QD optimization \cite{huber2024speeding}. For each placement pose, the generated grasps are evaluated in a simulation to guarantee their validity within the placing setup. This process yields a set of grasps, corresponding placement poses, and associated quality metrics—including whether $X_i^o(T)$ is a stable equilibrium with respect to $\sigma^{DR}$. The object is then grasped and placed using the corresponding poses and a motion planner.

%% file: tex_files/figures/qd_place_principle.tex
\begin{figure}[t]
  \centering
\centering
  \includegraphics[width=0.9\columnwidth]{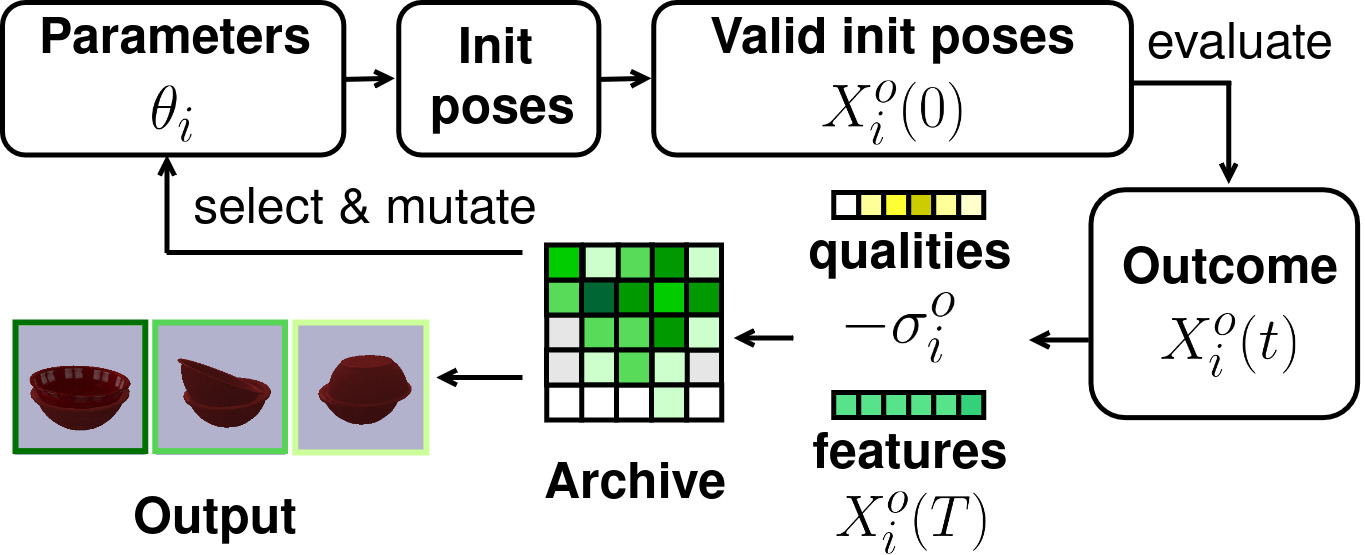}
  \caption{\textbf{Placeit! principle.} The framework uses QD optimization to explore the interactions between an object and a support. A mutation-selection process efficiently explores the space of possible placement solutions from a parameter space.}
  \label{fig:qdp_principle}
\end{figure}

%% file: tex_files/figures/qdgp_principle.tex
\begin{figure}[t]
  \centering
\centering
  \includegraphics[width=0.8\columnwidth]{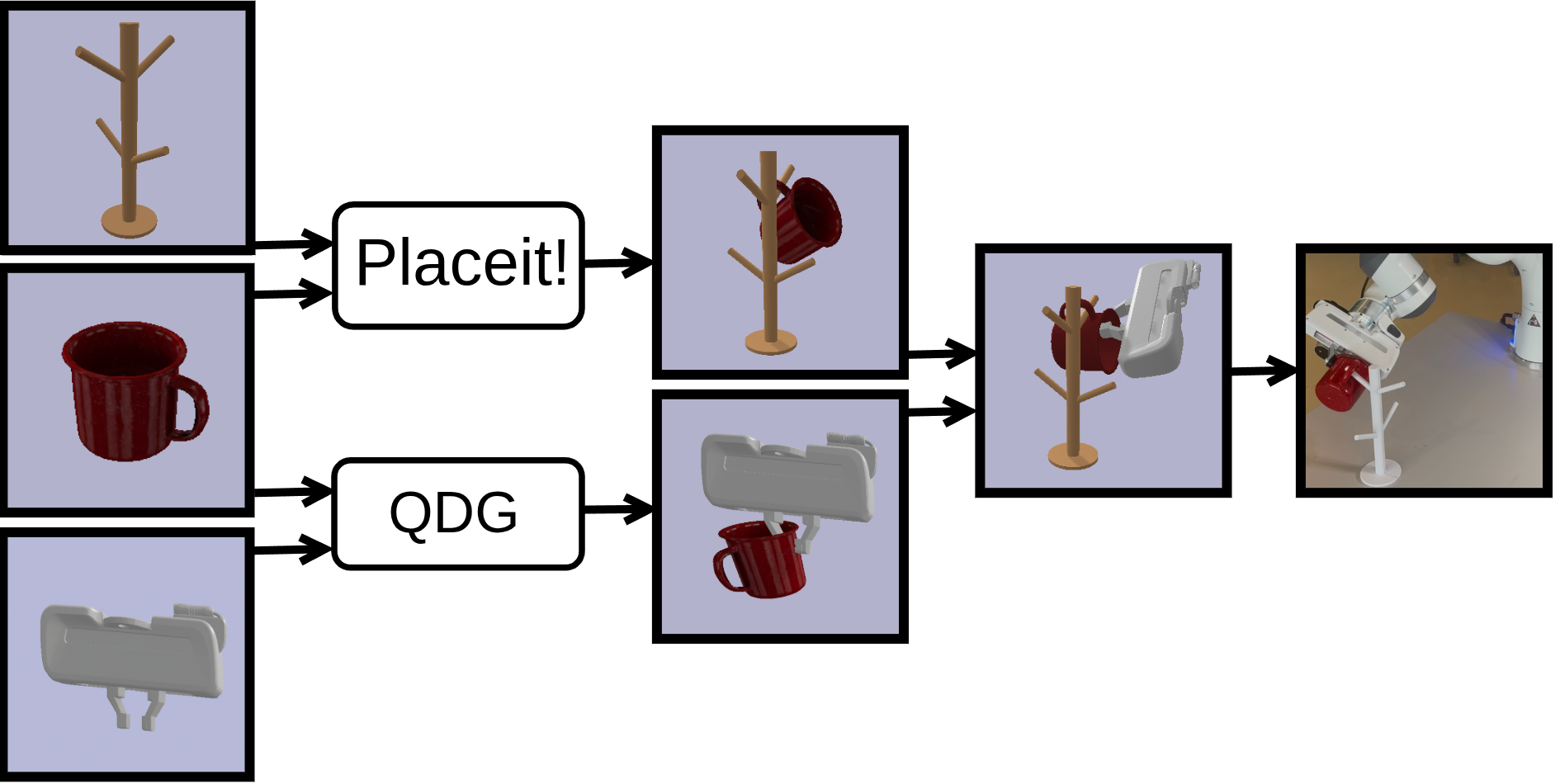}
  \caption{\textbf{QDGP pipeline.} Grasp poses are generated using QDG \cite{huber2024speeding}, and placing poses with Placeit!. The combined poses are validated before being deployed on the physical robot to perform the full pick-and-place sequence.
  }
  \label{fig:qdgp_principle}
\end{figure}

%% file: tex_files/figures/qdplace_tasks.tex
\begin{figure}[t]
    \centering
    \includegraphics[width=\columnwidth]{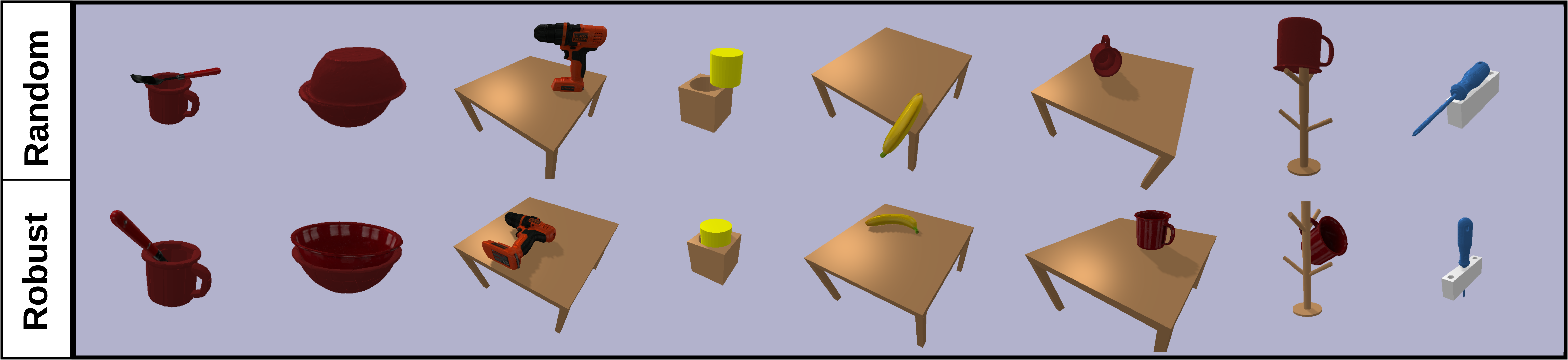}
    \caption{\textbf{Example solutions.} The top row shows solutions randomly sampled from a CMA\_MAE run. The bottom row shows robust solutions from the same run, identified using our domain randomization criterion.}
    \label{fig:considered_task}
\end{figure}

%% file: tex_files/4_experiments.tex
\input{tex_files/figures/exp1_results}

\section{EXPERIMENTS} 
\label{sec:experiments}

This section presents the essential information about our experiments. Further details, including hyperparameters, can be found in the Supplementary Materials.


\textbf{\textit{Tasks.}} An overview of the tasks we considered can be found in Fig. \ref{fig:considered_task}. We defined 8 scenarios in simulation to compare the placement sampling schemes. These scenarios included: 3 tabletop tasks (\textit{banana-on-table}, \textit{mug-on-table}, \textit{power drill-on-table}); 1 stacking task (\textit{stack-bowl}); 3 insertion tasks (\textit{peg-in-hole}, \textit{spoon-in-mug}, \textit{screwdriver-on-support}); and one hanging scenario (\textit{hanging-mug}). These tasks were chosen to demonstrate the framework's versatility across various placement challenges, reflecting different levels of complexity. They also correspond to scenarios proposed in works used as baselines \cite{harada2014validating, zhao2025anyplace, noh2024learning, simeonov2023shelving}.



\textbf{\textit{Compared methods.}} Our experiments evaluate two critical components of the Placeit! framework: the search space and the optimization algorithm. We structured the comparison into four main parts: state-of-the-art baselines, QD approaches, search space baselines, and a final combination of a prior-based search space with QD.

The existing literature on generating placement poses primarily relies on three types of prior-based methods:

\begin{itemize}
    \item \textit{face\_alignment}: This approach involves randomly sampling and aligning a triangle mesh from both the object to be placed and the support object \cite{harada2014validating, jermann2024efficient, street2024right}. A random orientation is applied, and the validity is checked by simulating gravity;
    \item \textit{pca\_alignment}: This method uses Principal Component Analysis (PCA) to align the two objects based on their symmetrical axes. A random perturbation is then applied, and the object is dropped to check its validity \cite{zhao2025anyplace};
    \item \textit{rand\_sample}: This approach samples a random initial pose for the object relative to the support, then applies gravity to see if it settles into a stable state \cite{noh2024learning}.
\end{itemize}

We also included several Quality-Diversity methods, which utilize a naive, prior-less search space where the object's position and orientation are directly controlled by the algorithm's genome. The object is then "dropped" to test its stability. The QD variants we included are:

\begin{itemize}
    \item \textit{ME\_rand}: A standard MAP-Elites baseline \cite{mouret2015mapelites};
    \item \textit{ME\_scs}: A success-greedy variant of MAP-Elites, known for its efficiency in tasks like grasping \cite{huber2023quality};
    \item \textit{CMA\_MAE}: A state-of-the-art QD approach that has performed well across many domains \cite{fontaine2023cmamae}.
\end{itemize}

To investigate the impact of reducing the search space, which has been shown to improve QD performance \cite{huber2024speeding}, we introduced a prior-based search space called \textit{contact\_rand\_sample}. This approach randomly samples an initial state for the object relative to the support object's surface. A triangle mesh on the support object is selected, and the object to be placed is positioned and oriented within a cone around the triangle's normal—similarly to what Eppner et al. did for grasping \cite{eppner2023abw2g}. We also created combinations of the prior-based search space with each of the QD methods (\textit{contact\_ME\_rand}, \textit{contact\_ME\_scs}, and \textit{contact\_CMA\_MAE}) to evaluate their combined performance.



\textbf{\textit{Pick\&Place in the real-world.}} To evaluate the quality of the placement poses found in simulation and their applicability in the real world, we conducted a sim-to-real transfer experiment. A set of 7 diverse scenarios were chosen, including tabletop, insertion, stacking, and hanging tasks. The scenarios used in this transfer experiment differ slightly from those in our main simulated experiments due to the accessibility of corresponding physical objects (e.g., the stack-bowl scenario was replaced with a bowl-on-plate). We performed dedicated runs of Placeit! for these specific scenarios to generate the poses for real-world testing. A quality label based domain randomization (eq. \ref{eq:dr_validation_place}) is also computed to distinguish robust poses from the ones at an unstable equilibrium.

Experiments were performed using a Franka Research 3 robotic arm equipped with its native two-finger gripper. The objects primarily came from the YCB dataset \cite{calli2015benchmarking}, but some were 3D-printed, such as a custom peg-in-hole task and a hanger for the mug to match a baseline study by Zhao et al. \cite{zhao2025anyplace}. QDGP relies on MoveIt as a motion planner.


\textbf{\textit{Evaluation metrics.}} The simulation experiments aim to find the best approach for generating a placing position in any of the considered scenarios. To do so, we measure the coverage of $\Phi^s$, noted $cvg(\Phi^s)$, which is the ratio of successful placing states relative to all the theoretically possible placing states in $\Phi$. To evaluate the quality of the generated placing poses, we deploy them in the real world and measure whether the attempt is successful or not. A placing position is considered successful if the object reaches a stable state in the real world that is similar to the one found in simulation. Therefore, a placing attempt that accidentally leads to another valid placing state in the real world is not considered a success.


%% file: tex_files/figures/exp1_results.tex
\begin{figure*}[thpb]
  \centering
  \includegraphics[width=\textwidth]{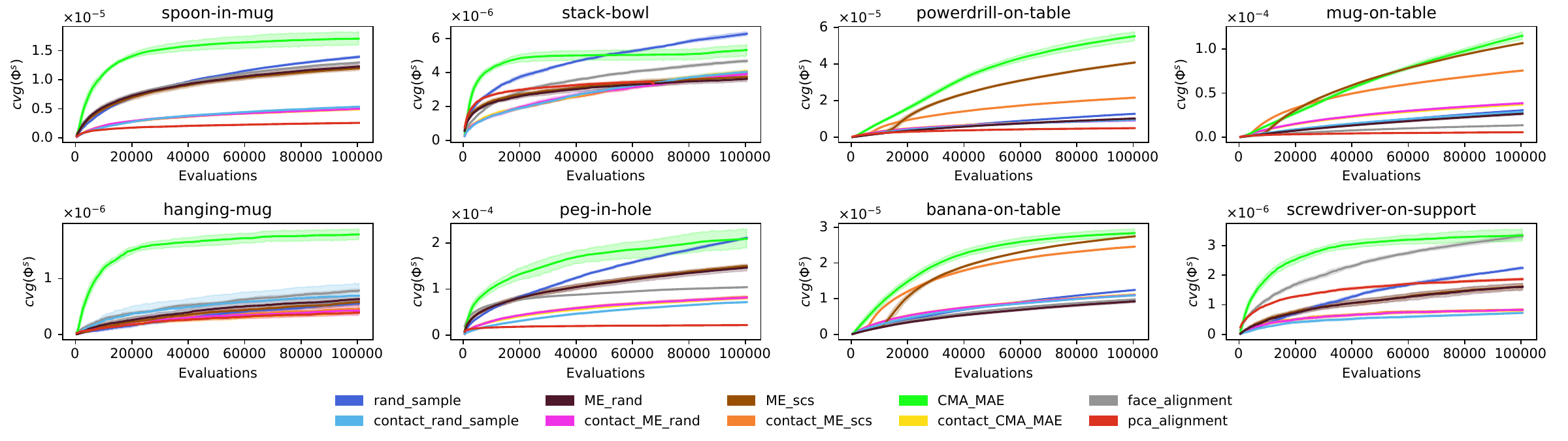}
  \caption{\textbf{Comparison between methods.} This figure shows the evolution of the diversity of placement solutions found ($cvg(\Phi^s)$) throughout the optimization process for each simulation scenario. The CMA\_MAE method without priors significantly outperforms all other methods across most scenarios.}
  \label{fig:exp1_results}
\end{figure*}

%% file: tex_files/5_results_and_discussion.tex

\section{RESULTS AND DISCUSSION}
\label{sec:results_and_discussion}

\input{tex_files/figures/placing_heatmaps}

\input{tex_files/figures/variance_to_perturbation_distribution}


\textbf{\textit{Placing in Simulation.}} As shown in Fig. \ref{fig:exp1_results}, our experiment compares the performance of various methods. The QD state-of-the-art method without any prior (CMA\_MAE) significantly outperforms all other approaches, particularly in the short run ($<$ 20k evaluations), where it excels across all scenarios. While random sampling (rand\_sample) can eventually generate a greater diversity of poses in the long run (100k evaluations), much of this diversity is often redundant or fragile. CMA\_MAE, by contrast, generates both diverse and high-performing solutions efficiently, avoiding less promising areas of the search space. This difference explains why random sampling can become competitive with CMA\_MAE in specific long-run scenarios, such as rotations around the symmetrical axis in stack-bowl or peg-in-hole.

State-of-the-art methods relying on alignment priors (face\_alignment, pca\_alignment) are consistently outperformed by most of our QD variants. Although these priors can be effective in specific, symmetric scenarios (e.g., screwdriver-on-support, stack-bowl) or for insertion-like tasks (e.g., peg-in-hole, hanging-mug), their performance is highly task-dependent. In contrast, our generalist QD approach (CMA\_MAE) consistently provides superior sample efficiency, regardless of the scenario.

Our experiment also reveals that reducing the search space by forcing close contact between objects (contact\_*) does not improve search efficiency for placing tasks. This finding contradicts its effectiveness in grasping scenarios \cite{huber2024speeding}. We attribute this to the sparsity of the problem. Placing is a less sparse task than grasping, meaning a higher ratio of attempts yield successful solutions, which has a notable impact on QD methods' performance \cite{huber2023quality}. This also explains why CMA\_MAE dominates ME\_scs on most tasks. While ME\_scs excels in problems with very low success rates and a concentrated region of interest \cite{huber2023quality}, CMA\_MAE performs best when it can rely on more frequent exploitable signals to balance exploration and exploitation \cite{fontaine2023cmamae}. This result highlights the need for a better understanding of the link between QD optimization algorithms and task sparsity, which is a promising direction for future work.

\input{tex_files/figures/real_place}

\input{tex_files/tables/sim2real_deployment_results}


\textbf{\textit{Interaction properties.}} Since the best-performing methods in our study thoroughly explore how two objects interact, we can infer physical properties from the resulting set of object states. For instance, our framework can identify the parts of an object most likely to contact a support for a given scenario, as seen in Fig. \ref{fig:placeing_heatmap}. By creating a continuous map, we can pinpoint different placement modalities. When plotting the distribution of state variations after applying a local perturbation from a given placement pose (Fig. \ref{fig:variance_to_perturbation_distribution}), the resulting modes correspond to the number of ways the objects can be placed on the support. Thus, the Placeit! framework can automatically identify how objects can be placed on one another, which allows for a broader analysis of object similarities and manipulation invariants. For example, all objects with a handle-like subpart could be automatically identified as something that can be hung in various ways—a property that would have emerged from prior interactions of similar objects with hangers.

This approach also reveals properties related to the simulation environment itself. For example, Fig. \ref{fig:variance_to_perturbation_distribution} shows that in Bullet, placing the YCB mug in its common-use pose (*) is more stable than flipping it (***). This is counterintuitive, as in the physical world, placing a mug on its top or bottom face typically yields comparable stability, both being less stable than on its side (**). In the simulation, the modeling of contacts between the 3D-scanned object and the surface causes the object state to vary significantly more when placed in pose (***) than in pose (*) or (**). This is another instance where domain randomization can stress the limits of simulations and be used as a tool for analyzing and improving both physics engines and rigid body meshes \cite{huber2023domain}.


\textbf{\textit{Pick\&Place in the real-world.}} Table \ref{table:s2r_deploy_results} shows the successful sim-to-real transfer ratios achieved using QDGP across various placing scenarios. In every case, sampling poses that satisfy the stable equilibrium criterion leads to a higher transfer ratio compared to random sampling. While both methods perform well on easier scenarios like mug-on-table and bowl-on-plate, the difference is far more significant for tasks that require precision, such as peg-in-hole, screwdriver-on-support, and hanging-mug. This demonstrates that our domain randomization approach is effective at identifying which simulated placement poses are most likely to succeed in the real world by distinguishing local or global extrema of potential energy.

Fig. \ref{fig:sim2real_place_vis} shows several examples of successful deployments. The Placeit! approach led to a high success rate in placing manipulations for all scenarios, proving its utility for direct deployment in both industrial environments (e.g., screwdriver-on-support) and service robotics (e.g., bowl-on-plate). The demonstrated effectiveness of this approach in real-world scenarios confirms its potential for broader applications, such as learning to manipulate objects on-the-fly by scanning an environment \cite{hampali2023inhand} or generating labels for learning placing policies \cite{simeonov2023shelving}. The framework's versatility, combined with the sample efficiency of QD optimization and the quality of its sim-to-real transfer criteria, makes it a promising tool for making future manipulation task learning easier and more efficient.


\textit{\textbf{Limitations.}} This work focuses on the placement phase of object manipulation, assuming the preceding approach can be handled with standard planning. This assumption may not hold true in dense scenes \cite{haustein2019object}. We believe the data generated by our framework unlocks a critical part of the placing skill, and future work could explore how to integrate these placements with dedicated methods, such as reinforcement learning, to address this challenge. Furthermore, our experiments were limited to rigid, non-deformable objects. While the Placeit! framework is versatile enough to handle other types, extending our experimental results to include deformable or articulated objects is a promising direction for future research. Finally, Placeit!'s performance is highly dependent on the quality of its simulation environment, since it relies entirely on it. The success of our approach is critically tied to the quality of the object meshes, the accuracy of their inertial properties, and the precise modeling of contacts within the physics engine.

%% file: tex_files/figures/placing_heatmaps.tex
\begin{figure}[t]
  \centering
\centering
  \includegraphics[width=\columnwidth]{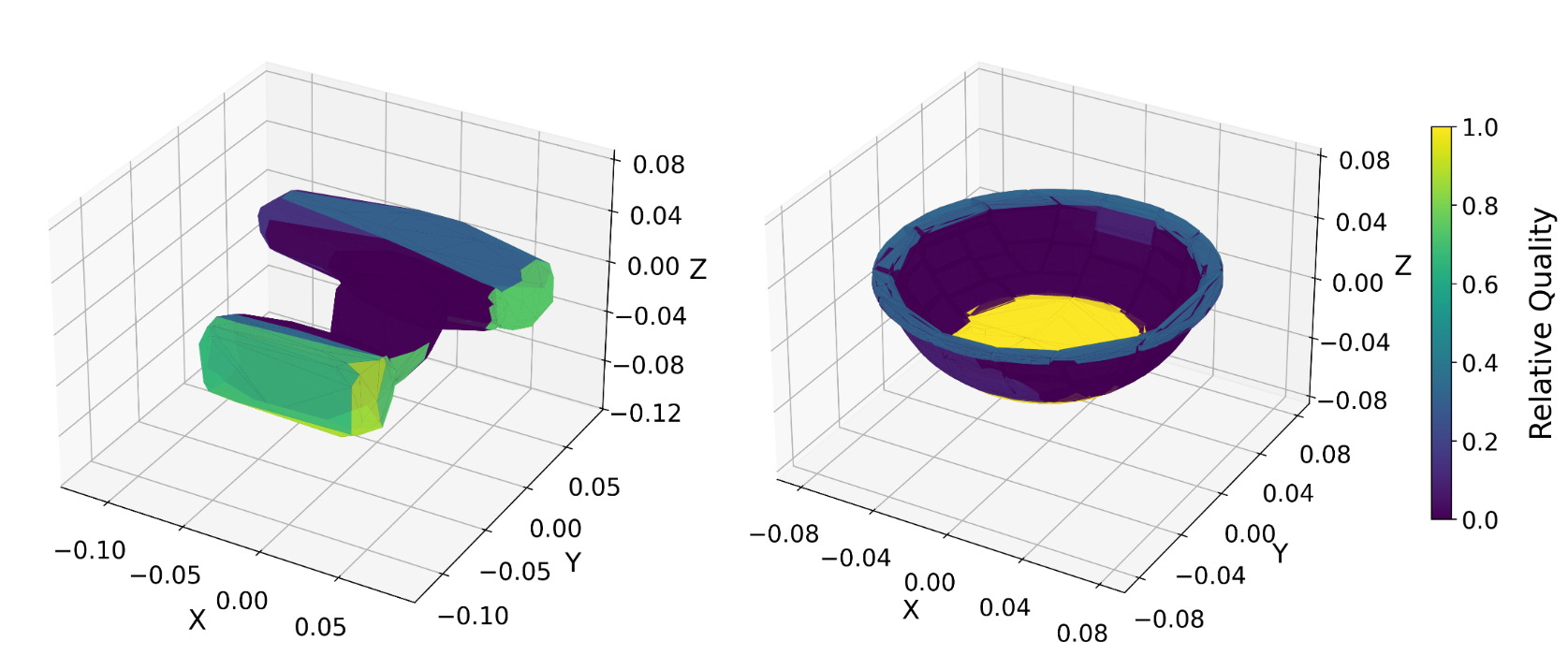}
  \caption{\textbf{Placing heatmaps for 2 table-top scenarios.} The heatmaps show the relative quality of placement contacts. Areas of the mesh that are most often in contact with the table have a higher value. (See supplementary materials for details.)}
  \label{fig:placeing_heatmap}
\end{figure}

%% file: tex_files/figures/variance_to_perturbation_distribution.tex
\begin{figure}[t]
  \centering
\centering
  \includegraphics[width=\columnwidth]{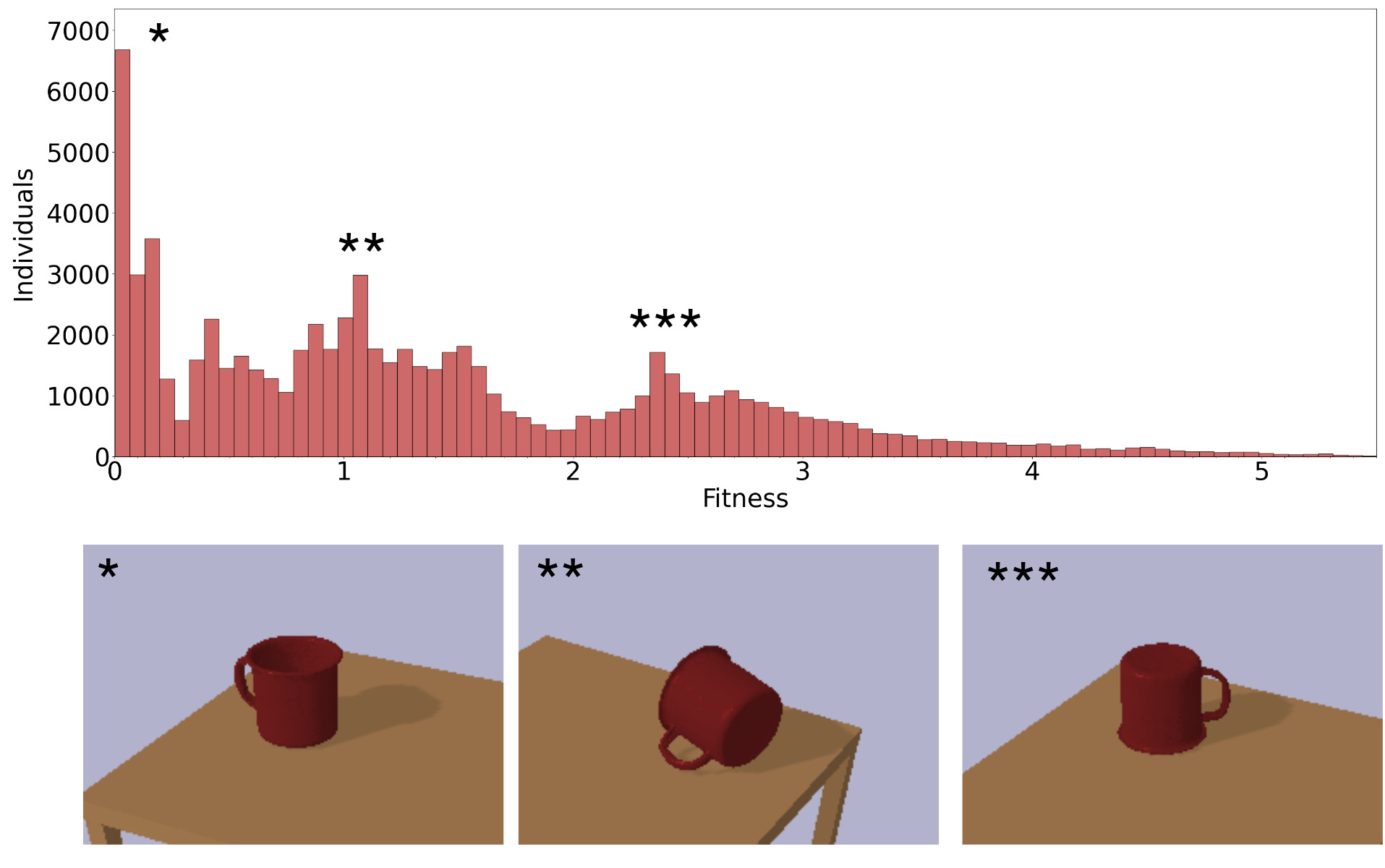}
  \caption{\textbf{Resistance to local perturbations.} The histogram shows the distribution of placing poses based on their pose variance after a local perturbation (fitness). This data, computed from 10 seeds and averaged over 10 random perturbations per pose, reveals distinct modes corresponding to different stable ways for objects $o$ and $s$ to interact. The visualizations are randomly sampled poses from each of these peaks. This example highlights a limitation of the used simulator: placing the mug on its side (*) is much more stable than placing it on its flipped side (***).
  }
  \label{fig:variance_to_perturbation_distribution}
\end{figure}

%% file: tex_files/figures/real_place.tex
\begin{figure}[t]
  \centering
  \includegraphics[width=\columnwidth]{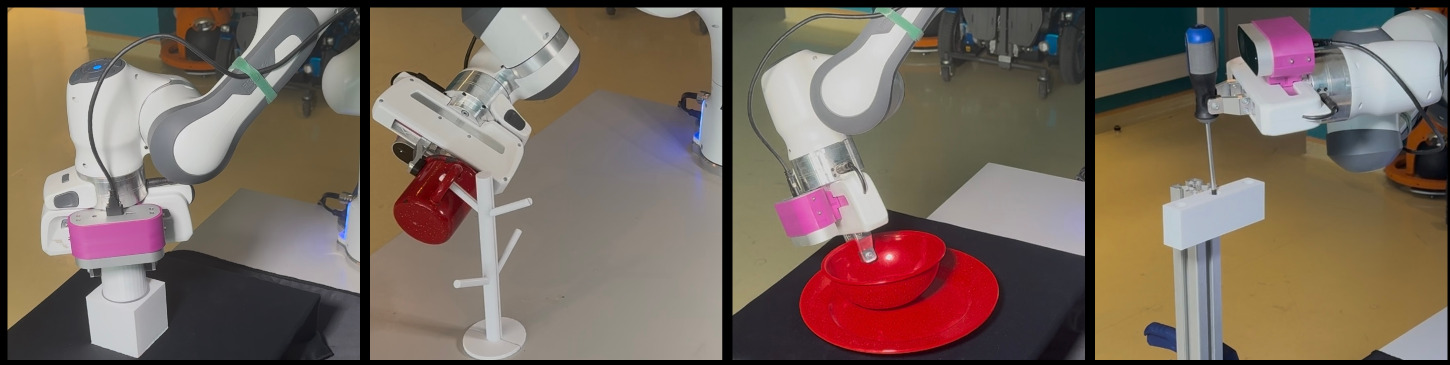}
  \caption{\textbf{Real-world transfer.} The proposed framework produces diverse placement poses that achieve a high success rate when deployed in the physical world. This demonstrates the reliability of our approach for robotic manipulation learning.}
  \label{fig:sim2real_place_vis}
\end{figure}


%% file: tex_files/tables/sim2real_deployment_results.tex
\begin{table}[t]
\centering
\begin{tabular}{ ||c || c | c ||}
\hline
Scenarios & Random & Robust  \\
\hline
\hline
banana-on-table & 0.90 & 1.00 \\
\hline
mug-on-table & 1.00 & 1.00 \\
\hline
peg-in-hole & 0.56 & 0.78 \\
\hline
spoon-in-mug & 0.60 & 0.90 \\
\hline
bowl-on-plate & 0.90 & 1.00 \\
\hline
screwdriver-on-support & 0.40 & 0.60 \\
\hline
hanging-mug & 0.00 & 1.00 \\
\hline
\hline
Total & 0.62 & 0.90 \\
\hline

\end{tabular}
\caption{\textbf{Real world transfer ratios for randomly sampled \\  placing poses and for poses in stable equilibrium.}}
\label{table:s2r_deploy_results}
\end{table}

%% file: tex_files/6_conclusions.tex
\section{CONCLUSIONS}

In this work, we introduce Placeit!, a framework for the automatic generation of rigid object placement poses in simulation. Our method is versatile, applying to various scenarios like table-top, hanging, stacking, and insertion. We demonstrate that a Placeit!-based approach to pick-and-place achieves a 90\% success rate, proving the framework's efficiency in teaching robots manipulation skills for known objects in open environments. Furthermore, the high quality of the generated synthetic data suggests that Placeit! is a highly promising tool for building future foundation models in robotics.

%% file: tex_files/acknowledgment.tex

\section*{ACKNOWLEDGMENT}

This work was supported by the German Ministry of Education and Research (BMBF) (01IS21080), the French Agence Nationale de la Recherche (ANR) (ANR-21-FAI1-0004) (Learn2Grasp), the European Commission's Horizon Europe Framework Programme under grant No 101070381 and from the European Union's Horizon Europe Framework Programme under grant agreement No 101070596.

%% file: tex_files/appendices.tex

\appendices

\begin{center}
\textbf{\large Supplementary Materials}
\end{center}


\section{Experimental details: simulation}
\label{sec:a1_exp_details_sim}


\textbf{Algorithms.} Let $\mu$ be the population size, $\lambda$ the number of offspring, $k$ the number of neighbors considered for novelty computation, and $N_e$ the maximum number of evaluation. We set: $\mu=\lambda=500$, $k=15$, $N_e=100\text{k}$. All offspring are mutated with a probability $ind_{pb}=0.3$ to modify each gene. The mutation operator applied by default to all the methods is a Gaussian perturbation of $0$ mean and $0.1$ standard deviation. For \textit{CMA\_MAE} variants, the same parameters as in Fontaine et al. were used \cite{fontaine2023cmamae}: The emitter batch size is set to $36$, and the number of emitters to $15$ and $\alpha=0.2$. Finally, we set $f_{min} = -10000$ to make sure $f_i>f_{min})$.


\textbf{Heatmaps.} The heatmaps in Fig. \ref{fig:placeing_heatmap} were computed based on a \textit{relative quality}: for a given archive of successful pose, we extract a chunk of object point cloud of 10\% along the z axis from the table surface. Each chunk is an approximation of the points on the object surface that are in contact with the table. The obtained points are associated with the computed fitness for each chunk. All of the collected fitnesses for each points are then summed. The relative quality for a given point is computed as the ratio between its associated sum of fitnesses and the maximum value among all the sums. This allows to obtain a score between 0 and 1, where scores close to 1 emphasize points that are often in contact with the table and associated with a high value of fitness. This computation is similar to the QD-score, which is commonly used to encapsulate both the diversity and the quality in a single value \cite{cully2022qd}.


\section{Experimental details: physical world}
\label{sec:a2_exp_details_real}


\textbf{Data generation.} To compute $\sigma_i^{o^{DR}}$, an axis of perturbation (among the translation and the rotation ones) is first randomly sampled. The magnitude of the perturbation depends on the mass of the object, such that the perturbations on each object are comparable.


All the remaining hyperparameters are similar to those used in the simulated experiments.


\textbf{Real world setup.} The gripper was controlled with a joint impedance controller with gravity compensation. No additional material that could increase adhesion is used to make grasping easier. The gravity compensation is computed with respect to the fixed arm base. We used a custom vision code derived from \cite{helenon2023learning} for getting the position and orientations of the objects $o$ and $s$.  The motion planning of the end effector is conducted with RRT connect through \textit{Moveit!}, assuming that the object and the table are collision bodies.

